%% file: main.tex
\documentclass[lettersize,journal]{IEEEtran}
\usepackage{amsmath,amsfonts}
\usepackage{array}
\usepackage[caption=false,font=normalsize,labelfont=sf,textfont=sf]{subfig}
\usepackage{textcomp}
\usepackage{url}
\usepackage{verbatim}
\usepackage{graphicx}
\usepackage{cite}
\usepackage[utf8]{inputenc}
\usepackage[T1]{fontenc}
\usepackage{graphicx}
\usepackage{multirow}
\usepackage{booktabs}
\usepackage{xcolor}
\usepackage{dirtytalk}
\usepackage{comment}
\usepackage{caption}
\usepackage{array}
\usepackage{dblfloatfix}
\usepackage{tabularx}
\usepackage[normalem]{ulem}
\usepackage{multirow}
\usepackage{rotating}
\usepackage{algpseudocode}
\usepackage{todonotes}
\usepackage{longtable}
\usepackage{adjustbox}
\usepackage{float}
\usepackage{tikz-dependency}
\usepackage{amssymb}
\usepackage{pifont}
\usepackage{tabularx}
\usepackage[many]{tcolorbox}
\usepackage{hyperref}
\usepackage{soul}
\newcommand*\circled[1]{\tikz[baseline=(char.base)]{        
 \node[shape=circle,fill,inner sep=0.5pt] (char) {\textcolor{white}{#1}};}}

\newcommand{\ie}{\textit{i.e.,}}
\newcommand{\eg}{\textit{e.g.,}}
\newcommand{\etal}{\textit{et al.}}

\newcommand{\findingsbox}[1]{
\begin{tcolorbox}[breakable,width=\linewidth,
boxrule=0pt, leftrule = 6pt, top=1pt, bottom=1pt, left=1pt,right=1pt, 
colback=gray!20,colframe=gray!60]
\textbf{Summary:} #1
\end{tcolorbox}
}

\iftrue           
 \newcommand{\MonzurulComment}[1]{\textcolor{purple}{#1}}
 \else
 \newcommand{\MonzurulComment}[1]{}
 \fi

\begin{document}

\title{ On the Sustainability of AI Inferences in the Edge }
\author{
 \IEEEauthorblockN{Ghazal Sobhani, Md. Monzurul Amin Ifath, Tushar Sharma, Israat Haque} \\
\IEEEauthorblockA{Dalhousie University}
}

\maketitle

\begin{abstract}

The proliferation of the Internet of Things (IoT) and its cutting-edge AI-enabled applications (e.g., autonomous vehicles and smart industries) combine two paradigms: data-driven systems and their deployment on the edge. Usually, edge devices perform inferences to support latency-critical applications. In addition to the performance of these resource-constrained edge devices, their energy usage is a critical factor in adopting and deploying edge applications. Examples of such devices include Raspberry Pi (RPi), Intel Neural Compute Stick (INCS), NVIDIA Jetson nano (NJn), and Google Coral USB (GCU). Despite their adoption in edge deployment for AI inferences, there is no study on their performance and energy usage for informed decision-making on the device and model selection to meet the demands of applications.    
This study fills the gap by rigorously characterizing the performance of traditional, neural networks, and large language models on the above-edge devices. Specifically, we analyze trade-offs among model F1 score, inference time, inference power, and memory usage.
Hardware and framework optimization, along with external parameter tuning of AI models, can balance between model performance and resource usage to realize practical edge AI deployments.

\end{abstract}


\section{Introduction}
\input{01_introduction} \label{intro}

\section{Background}
\input{02_background_motivation}

\section{Related work}
\input{03_related_work}

\section{Methodology}
\input{04_methodology}

\section{Implementation and Deployment}\label{sec:implementation}
\input{05_implementation}
\section{Evaluation Results}
\input{06_evaluation}

\input{07_discussion}

\section{Conclusion}

\input{08_conclusion}

\bibliographystyle{IEEEtran}
\bibliography{references}
\end{document}

%% file: 01_introduction.tex
Edge Artificial Intelligence (AI) represents a paradigm shift from traditional cloud-based AI, which centralizes computation in data centers, towards processing on edge devices such as IoT sensors, edge servers, and industrial equipment. This proximity to the data source facilitates real-time decision-making and inference directly on the edge devices.
Edge AI is gaining prominence due to several factors, including reduced latency, enhanced privacy and security, improved bandwidth efficiency, and the decentralization of computation~\cite{zhang2018, wang2020}.
By processing data locally, Edge AI significantly reduces latency, which is critical for applications such as autonomous vehicles, robotics, and industrial automation. 
Edge AI enhances privacy and security by keeping sensitive data on-premises, minimizing the need to transmit data to external servers.
This on-device computation reduces bandwidth usage by minimizing large-scale data transfers. Additionally, distributing processing across edge devices improves system resilience and responsiveness through decentralized computation.

Recent advancements in hardware platforms, such as \textit{System on Chips} (\textsc{SoCs}), \textit{Field Programmable Gate Arrays} (\textsc{FPGAs}), 
and \textit{AI accelerators}, have made it feasible to run inference workloads on resource-constrained devices \cite{mao2024green,huang2025edgellm}. However, the widespread deployment of such devices, from Raspberry Pi to Neural Compute Stick and Google Coral, has raised concerns about the energy efficiency and sustainability of AI inference. With billions of AI-enabled edge devices projected to be deployed globally, the aggregate energy footprint of model inference is becoming a pressing concern. Green AI \cite{Georgiou2022} has emerged as a critical approach within artificial intelligence, emphasizing energy efficiency and sustainability.

Understanding the power consumption of ML and DL models is crucial for optimizing their deployment in edge computing environments. By quantifying the power requirements of ML and DL models, developers can make informed decisions about model selection, optimization techniques, and hardware configurations. This knowledge is instrumental in developing energy-aware algorithms and strategies that minimize energy consumption without compromising model accuracy or performance. 
Ultimately, an awareness of energy consumption paves the way for designing sustainable edge computing systems, 
reducing operational costs and environmental impact. Furthermore, by prioritizing energy efficiency throughout the development and deployment process, organizations can contribute to a future of sustainable AI solutions that align with both their sustainability goals and regulatory requirements.



Current research on Edge AI often focuses on specific models or hardware, with limited attention to comprehensive performance and power measurement. There has been a few attempts in assessing the performance or the power usage of hardware devices. 
For example, Rafal \etal{}~\cite{Rafal2023} compared deep learning models (e.g., CNN-based models) on platforms like NVIDIA Jetson Nano and Intel Neural Stick, focusing mainly on inference speed. Similarly, Zhang \etal{}~\cite{zhang2018} analyzed machine learning frameworks to draw insights related to computing resources. Georgiou \etal{}~\cite{Georgiou2022} compared various identical implementations written in PyTorch and TensorFlow to measure their power consumption for large models. However, none of these measurements provided a comprehensive evaluation of performance and resource trade-offs across diverse models and edge devices for informed decision making in Edge AI deployments. 

This work fills that gap by conducting an in depth investigation of performance and resource tradeoff across various AI models and hardware targets. 
Various edge hardware supports distinct AI frameworks (e.g., LiteRT for Rasberry Pi, TensorRT for Jetson, OpenVINO for Intel, and EdgeTPU for Coral); this diversity creates fragmentation in deployment and measurement strategies. These inconsistencies present a challenge in establishing a standardized, fair, and reproducible evaluation methodology for edge AI. This paper addresses these gaps by proposing a unified inference assessment scheme by answering the following research questions.

\begin{description}
    \item[\textbf{RQ1.}] 
    \textit{Does domain-specific hardware and software design boost edge AI?}
    \item[\textbf{RQ2.}]
    \textit{What are the trade-offs between performance and resource usage in edge AI deployments?}
    \item[\textbf{RQ3.}]
    \textit{Which parameters can further be tuned for better power consumption in deep and large models?}
\end{description}

We extensively evaluated the proposed measurement scheme over 
 four widely used edge devices: Raspberry Pi, Intel Neural Compute Stick, NVIDIA Jetson Nano, and Google Coral. This study spans various AI model categories, including machine learning (ML), deep learning (DL), and large language models (LLMs). Additionally, we evaluated the impact of machine learning frameworks and various model parameters on the performance and resource usage of chosen models. 
 
We summarize our contributions (C) and observations (O) below.


\begin{itemize}
    \item [\textbf{C1:}] We propose a unified \textit{measurement approach} that integrates \textit{hardware-based power profiling} and \textit{software-based performance monitoring}, enabling consistent, reproducible evaluation across edge platforms. 

    \item [\textbf{C2:}]
    We offer a holistic view of edge device capabilities by conducting extensive performance evaluation of learning models, lightweight ML frameworks, and model parameters.

    \item [\textbf{C3:}] 
    We provide detailed guidelines for executing these models on edge devices, including best practices for optimizing model execution across different hardware configurations. 
        
    \item [\textbf{C4:}] 
    Our scripts, data, and experiment configurations are available in our replication package\footnote{To be released in GitHub \cite{edgeAI-gitRepo} upon paper acceptance.}.
    
    

    



    \item [\textbf{O1:}] Our results show that specifically designed edge devices can efficiently support different AI models (traditional to moderate) that can be further improved with optimized lighter AI frameworks. General purpose edge devices (\textit{e.g.,} Raspberry Pi) and frameworks (\textit{e.g.,} LiteRT) may offer better compatibility but may suffer from the performance and high resource usage.

    \item [\textbf{O2:}] 
    Carefully selecting edge device and framework is not enough to balance between performance and resource usage.  Inference parameter tuning plays a critical role in edge AI deployments.

    \item [\textbf{O3:}] 
    The proposed measurement scheme and evaluation findings will guide users to choose right combination of AI models, edge devices, frameworks, and parameters for successful edge AI deployments.

\end{itemize}

%% file: 02_background_motivation.tex

Integrating AI into edge computing environments, known as Edge AI, has revolutionized data processing. Traditionally, ML has empowered systems to learn and adapt from data without explicit programming to support applications such as predictive maintenance, anomaly detection, personalized healthcare, and industrial automation~\cite{mi2022, wang2020}. Natural language processing is also witnessing a paradigm shift with the potential deployment of Large Language Models (LLMs) on edge devices~\cite{Rahman2023}. This shift promises immediate language comprehension and generation, fostering applications in virtual assistants, language translation, and sentiment analysis. Deploying these models directly on edge devices facilitates real-time decision-making and analysis at the data source. However, deploying AI models on edge devices presents unique challenges, such as limited computational resources, energy constraints, and the need for lightweight model architectures~\cite{Rahman2023}. The following elaborates on the AI landscape, their conversions to lighter models, and resource measurement aspects on the edge. 

\textbf{Landscape of AI Models:}
AI models include classical machine learning models, neural networks, and large language models. Classical ML models, such as decision trees and support vector machines, are efficient and have a low-resource footprint, making them suitable for real-time tasks like predictive maintenance~\cite{wang2020}. Neural networks, particularly convolutional and recurrent neural network-based models, excel in complex pattern recognition for applications such as image recognition; however, these models require optimizations (\eg{} pruning) for edge deployment~\cite{Qi2018, wang2020}. LLMs, such as GPT, Llama, and Claude, offer advanced natural language processing capabilities but introduce challenges due to their large size and resource demands ~\cite{minaee2024}. Emerging techniques such as model distillation and federated learning~\cite{XIA2021100008} aim to enable the deployment of AI models in resource-constrained environments.

\textbf{Lightweight Model Conversions:}
To support these diverse AI models on constrained edge hardware, converting full-scale models into lightweight counterparts is essential. 
Techniques such as model pruning, quantization, and knowledge distillation help achieve compact models that maintain satisfactory accuracy levels~\cite{wang2020}. Quantization reduces memory usage and enhances computational efficiency by converting model weights to lower precision for faster inference times with acceptable accuracy levels~\cite{Rahman2023}. Additionally, techniques such as model pruning and knowledge distillation complement quantization by further streamlining models, making them ideal for real-time applications on edge devices~\cite{Rafal2023, wang2020}. 
Frameworks such as LiteRT~\cite{TensorFlowLite}, PyTorch Mobile~\cite{pytorch-mobile}, and ONNX Runtime~\cite{onnx} further empower the deployment of these models on edge devices. These frameworks enable tasks such as image classification, object detection, and speech recognition to be performed locally, eliminating the need for constant communication with centralized servers~\cite{zhang2018}. 

\textbf{Performance Measurement:}
The method of deploying and executing AI model to generate model's output on unseen data is referred as \textit{inference}.
To ensure the effectiveness of AI models deployed on edge devices, it is crucial to measure performance metrics such as inference speed, accuracy, memory utilization, and power consumption. \textit{Inference speed} indicates how quickly a model can process data, which is essential for real-time applications such as image recognition and natural language processing~\cite{Qi2018, wang2020}. \textit{Accuracy} assesses the model's ability to make correct predictions, ensuring reliability in critical tasks. 
\textit{Power consumption} measures the total power consumed to fulfill an inference request by a model. 

Software tools such as Perf~\cite{linux-perf} and NVIDIA-SMI~\cite{nvidia-smi}, along with hardware devices, USB power meters, provide valuable insights into the energy and power consumption of edge devices.
Performance can be measured using benchmarking tools provided by frameworks such as LiteRT and PyTorch Mobile~\cite{zhang2018}. These tools facilitate performance profiling by allowing developers to monitor resource usage and optimize models accordingly. By systematically evaluating these metrics, we can fine-tune model architectures and deployment strategies, ensuring efficient execution and responsiveness in edge applications.

\textbf{Why do we need edge AI measurements?}
The shift of inference workloads from cloud to edge devices has accelerated dramatically in recent years due to meet Quality of Service (QoS) of applications like autonomous vehicles, surveillance, and industrial automation. AI-enabled edge applications are common due to their effectiveness and access to adequate data. As edge inference continues to scale, encompassing traditional machine learning, lightweight neural networks, and even large language models, the cumulative energy footprint of these distributed computations is becoming substantial. 
For instance, recent study from Google indicates a substantial shift in AI-related energy consumption, with 60\% stemming from inference~\cite{patterson2022carbon}.
Despite this growing reliance, the landscape remains fragmented, characterized by diverse model architectures and deployment platforms but lacking standardized tools to measure power use in resource-constrained edge environments. Thus, this work focuses on 1) performing in-depth performance and power consumption benchmarking of edge devices with a wide range of AI/ML models and 2) developing a measurement scheme for informed decision-making in selecting and deploying learning-based edge applications with required QoS.

%% file: 03_related_work.tex

AI on edge devices have become pivotal for real-time AI applications; however, the resource-constrained characteristics of edge devices  pose significant challenges. Several studies have evaluated the performance of ML models on various hardware platforms. Rafal \etal{}~\cite{Rafal2023} assessed MobileNet, EfficientNet, VGG, and ResNet on platforms like NVIDIA Jetson Nano and Intel Neural Stick, showing that Google's hardware excels in inference times for newer architectures. Similarly, Zhang \etal{}\cite{zhang2018} compared ML frameworks TensorFlow, PyTorch, Caffe2, and MXNet, concluding that TensorFlow is ideal for large models on CPU-based systems, while PyTorch and MXNet offer better memory and energy efficiency.

Optimizing model deployment for edge devices is crucial for energy and runtime efficiency. Georgiou \etal{}~\cite{Georgiou2022} highlighted TensorFlow's advantage in recommender systems but noted PyTorch's superior runtime and energy efficiency in NLP tasks. Techniques like quantization play a key role in model deployment, as demonstrated by Rahman \etal{}~\cite{Rahman2023}, e.g., MobileBERT fine-tuned with TensorFlow Lite allows for efficient deployment on constrained hardware without significant accuracy loss. Frameworks such as TensorFlow Lite and Microsoft's EdgeML further enhance TinyML applications for IoT devices.

However, DL models on edge devices present several challenges. Qi \etal{}~\cite{Qi2018} discussed key optimization methods, including parallelization, quantization, and pruning, to balance model performance and resource limitations. Additionally, Ghosh \etal{}~\cite{Ghosh2019} proposed hybrid approaches combining edge and cloud computing for IoT data analytics, significantly reducing data transmission while maintaining task accuracy. May \etal{}~\cite{may2024dynasplit} jointly tuned model partitioning and hardware-level parameters under QoS constraints for edge-cloud inference optimization, using a two-phase design with offline profiling and online scheduling. Evaluated on models like VGG16 and ViT over Raspberry Pi and Coral (compiled with LiteRT), this method significantly reduced energy usage while meeting latency targets.

Recent work has begun to explore LLM inference on edge accelerators. Arya \etal{}~\cite{arya2025understanding} systematically benchmarked latest LLMs ranging from 2.7B to 32.8B parameters on Nvidia Jetson Orin, evaluating the effects of batch size, sequence length, quantization, and custom power modes. Their findings revealed trade-offs between throughput, latency, and energy under varying workloads, e.g., higher batch size improves throughput but also increases memory usage due to KV caching.
Overall, these studies underscore the growing need for optimized ML models and frameworks tailored for edge computing.

\begin{table*}[h!]
\centering
\caption{ A summary of the related work.}
\begin{tabular}{p{2.5cm}|p{6cm}|p{8cm}}
\textbf{Related Work} & \textbf{Contributions and gaps} & \textbf{Our contributions} \\ 
\toprule
Rafal \etal{}~\cite{Rafal2023} & 
Focuses primarily on CNN models for vision tasks, limiting model diversity. & 
Expanded model diversity by including a wider range of models, including large language models (LLMs), for broader applications. \\ 
\hline
Zhang \etal{}~\cite{zhang2018} & 
Limited performance metrics, primarily focusing on latency and memory usage. & 
Conducted a comprehensive evaluation with additional metrics such as inference time, memory consumption, and power usage for deeper insights. \\ 
\hline
Georgiou \etal{}~\cite{Georgiou2022} & 
Discusses energy efficiency but lacks detailed power consumption measurements across various models. & 
Performed detailed power consumption measurements, providing practical guidelines for power-efficient deployment on edge devices across multiple model types. \\ 
\hline
Rahman \etal{}~\cite{Rahman2023} & 
Covers quantization techniques but lacks broader model comparisons and application contexts. & 
Evaluated quantization across multiple models, emphasizing optimization strategies, deployment efficiency, and practical applications for TinyML. \\ 
\hline
Qi \etal{}~\cite{Qi2018} & 
Highlights optimization techniques but does not explore practical implementations or their impact on performance. & 
Explored practical implementations of optimization techniques, analyzing their effects on performance and resource limitations for edge devices. \\ 
\hline
Ghosh \etal{}~\cite{Ghosh2019} & 
Proposes hybrid approaches but lacks evaluation metrics for performance across edge and cloud systems. & 
Evaluated performance metrics for hybrid approaches, assessing accuracy and efficiency in real-world applications, particularly for IoT data analytics. \\
\hline
May \etal{}~\cite{may2024dynasplit} & 
Explores energy-aware NN partitioning across edge and cloud, but requires runtime coordination. & 
Focused on on-device inference and systematic energy evaluation without runtime dependencies. \\
\hline
Arya \etal{}~\cite{arya2025understanding} & 
Analyzes LLM inference on a single Jetson device but omits non-LLM models and cross-platform analysis. & 
Benchmarked diverse model types across multiple edge platforms under consistent power and performance evaluation. \\
\bottomrule
\end{tabular}
\label{table:comparison}
\end{table*}

\textbf{Comparison with Existing Work:}
The above evaluations fall short in their comprehension and systematic measurement scheme, which is the focus of this work. 
Specifically,  we assess a range of AI models, including large language models, enabling the deployment of a variety of edge applications. These models are deployed and tested across various edge devices for inference time, memory consumption, and power usage to offer deeper insights into the trade-offs involved in model optimization in edge deployment. We develop a systematic measurement scheme using hardware and software tools for fine-granular and dependable power and resource measurements. 
Table~\ref{table:comparison} provides a detailed comparison with related studies to showcase the gaps between the proposed and existing schemes.

%% file: 04_methodology.tex
\begin{figure}[h!]
\centering
\includegraphics[width=\columnwidth]{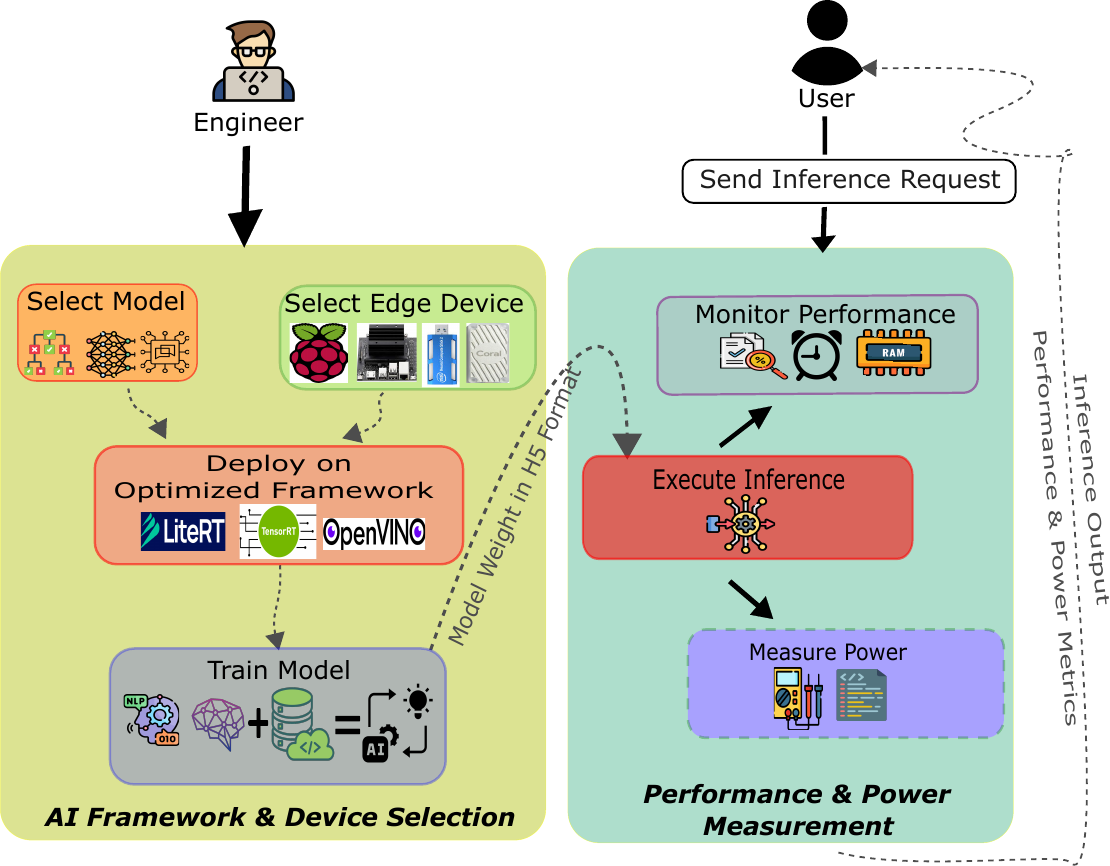}
\caption{The workflow of AI inference assessment in edge devices.}
\label{fig:edge_overview}
\end{figure}

This section details our measurement methodology by addressing the research questions outlined in Section~\ref{intro}. Figure~\ref{fig:edge_overview} illustrates the overall workflow for assessing AI inference in edge devices, encompassing two key phases: (i) selecting suitable AI frameworks and devices and (ii) developing a comprehensive performance and energy measurement scheme. 

\subsection{Selecting AI Framework for Edge Platforms (RQ1)}
\label{subsec:AI framework selection}

The framework selection consists of \circled{1} AI model selection to meet the edge application demands, \circled{2} edge platform selection, \circled{3} appropriate light AI framework selection, \circled{4} device-specific optimizations, and \circled{5} model training. We elaborate each step below.

\subsubsection{Model Selection}

We choose a set of models widely deployed in edge applications, e.g., image processing and classification, which are prevalent in AI applications on edge devices. Traditional ML and large language models are also used in real-time sensor data analysis and voice-activated systems, respectively. We ensure that the chosen models are compatible with the chosen devices and frameworks by aligning each category with relevant edge computing scenarios. This strategy thoroughly evaluates performance factors, such as accuracy, inference speed, memory utilization, and power consumption across various edge devices.

\begin{itemize}
\item \textit{Traditional ML Models:} These models rely on predefined mathematical formulations and are efficient in both computation and memory, making them suitable for lightweight edge applications.
Decision Trees, Linear Regressions, Support Vector Machines (SVM), and k-Nearest Neighbors (KNN) are effective in environments like analyzing real-time sensor data~\cite{wang2020}.

\item \textit{Neural Network Models:} Neural networks consist of layers of interconnected nodes that learn complex representations from data. They are well-suited for high-dimensional tasks such as image or speech processing.
Artificial Neural Networks (ANN), Convolutional Neural Networks (CNN), Region-based CNNs (R-CNN), ResNet-50, and MobileSSD are extensively utilized for image and video data processing. Some example applications include gesture recognition, autonomous driving, and video analysis~\cite{wang2020}. 

\item \textit{Large Language Models:} LLMs are deep learning architectures pre-trained on massive corpora to understand and generate natural language.
We select TinyBERT and Phi-2-orange models that are designed for tasks such as voice assistance and text analysis, demonstrating the capability of edge computing to handle diverse NLP applications~\cite{Rahman2023}.
\end{itemize}


\subsubsection{Device Selection}

We have chosen a diverse set of edge devices to accommodate the above mentioned AI models. Our list includes Raspberry Pi, Jetson Nano, Intel Neural Compute Stick, and Google Coral USB. The affordability and adaptability of the Raspberry Pi, the GPU capabilities of the Jetson Nano, and the inference acceleration provided by the Intel Neural Compute Stick and Google Coral USB together offer a well-rounded range of platforms for our experiments in edge computing. We elaborate on these devices below.

\begin{itemize}
\item \textit{Raspberry Pi (RPi):}
The RPi is well-regarded for its flexibility and cost-efficiency in edge computing scenarios~\cite{Rahman2023}. Its small form factor, low power usage, and General Purpose Input/Output (GPIO) ports facilitate easy integration with various sensors and peripherals. This makes it particularly suitable for IoT applications and educational projects~\cite{raspberrypi}. We use a Raspberry Pi 4 Model B with 4GB LPDDR4 RAM and a 1.5 GHz quad-core ARM Cortex-A57 CPU, running 64-bit Raspberry Pi OS.

\item \textit{Intel Neural Compute Stick (INCS):}
The Intel Neural Compute Stick is a compact USB device designed to enhance the performance of deep neural network inference on edge devices. Featuring the Intel Movidius Myriad X Vision Processing Unit (VPU) \cite{Movidius}, it delivers high-performance inference while maintaining low power consumption. This makes it ideal for real-time tasks such as image and speech recognition in edge computing environments~\cite{intel_stick}. We utilize Intel Neural Compute Stick 2 in our setup and connect it to RPi via USB 3.0. 

\item \textit{Google Coral USB (GCU) Accelerator:} It is also a compact, plug-and-play device designed to accelerate machine learning inference on edge devices. Powered by the Edge-Tensor Processing Unit (Edge-TPU), it enables fast and efficient execution of AI models, particularly for computer vision tasks. With low power consumption and compatibility with TensorFlow Lite models, the Coral USB is ideal for applications such as image classification, object detection, and real-time analytics in IoT and smart devices~\cite{coral_usb}. Likewise INCS, we use the Coral USB Accelerator interfaced with RPi over USB.

\item\textit{NVIDIA Jetson Nano (NJn):}
The NVIDIA Jetson Nano Developer Kit is equipped with a 128-core Maxwell GPU and a quadcore ARM Cortex-A57 CPU, along with 4GB LPDDR4 RAM, offering substantial computing power for edge AI applications~\cite{wang2020}. It supports major AI frameworks, including TensorFlow and PyTorch, making it appropriate for applications ranging from autonomous robotics to advanced surveillance systems~\cite{jetson}.
\end{itemize}

\subsubsection{Framework Selection}

After selecting the AI models and edge devices, we explore candidate AI frameworks LiteRT \cite{TensorFlowLite} (formerly knowns as TensorFlow Lite), PyTorch Mobile \cite{pytorch-mobile}, and MXNet \cite{mxnet} as potential edge computing platforms owing to their broad acceptance in academic and industrial settings \cite{zhang2018}. MXNet is recognized for its scalability and efficiency, particularly in distributed training and inference across multiple GPUs. While its strengths are evident in large-scale training tasks and scenarios requiring significant parallelism, these features may be less relevant for edge deployment. PyTorch Mobile offers seamless integration with PyTorch, featuring dynamic computation graphs that provide flexibility for edge applications. Despite its advantages, we have chosen LiteRT as the primary platform for our assessment for the following reasons.


\begin{itemize}
    \item \textit{Reliable and Established Deployment}: TensorFlow and TensorFlow Lite are known for their reliable and stable deployment capabilities, widely utilized in both research and industry settings.

    \item \textit{Efficient Training and Inference:} TensorFlow supports and optimizes training a wide range of AI models, including the ones we chose. Also, TensorFlow Lite is designed for efficient inference through advanced optimizations such as quantization and GPU acceleration~\cite{tf-quantization}.
    
    \item \textit{Extensive Platform Compatibility}: TensorFlow Lite supports various platforms, including ARM-based devices, Android, iOS, and microcontrollers, making it adaptable for multiple hardware environments.
    
    \item \textit{Comprehensive Deployment Utilities}: TensorFlow Lite offers straightforward model conversion and deployment processes via tools such as the TensorFlow Lite Converter, facilitating an efficient transition from model training to deployment~\cite{zhang2018, Qi2018}.
\end{itemize}


\subsubsection{Device-specific optimizations}

Different edge devices utilize distinct, target-specific optimizations for lighter frameworks. For instance, the NVIDIA Jetson platform employs \textit{TensorRT}~\cite{TensorRT}, a specialized SDK from NVIDIA designed to enhance deep learning model inference on NVIDIA GPUs. \textit{TensorRT} leverages various advanced techniques, including precision calibration (\eg{} FP16 and INT8 quantization), layer fusion, and kernel auto-tuning, to accelerate model inference. These optimizations lead to substantial improvements in both performance and efficiency while reducing latency.

In contrast, the Intel Neural Compute Stick relies on the Intermediate Representation (IR) format~\cite{IR}, which is a key feature of the Intel OpenVINO toolkit. The \textit{IR} format comprises an XML file that outlines the network architecture and a binary file that holds the model weights. \textit{IR} models are optimized for Intel hardware using techniques such as weight pruning, quantization, and operator fusion. These enhancements facilitate efficient inference on devices like the Neural Compute Stick. By employing hardware-specific optimizations like \textit{TensorRT} for NVIDIA Jetson and the \textit{IR} format for Intel devices, we ensure that our AI models are executed with maximum efficiency and effectiveness on the respective target hardware.

Similarly, the Google Coral USB Accelerator employs the \textit{EdgeTPU}, a custom-designed ASIC from Google that accelerates TensorFlow Lite models optimized for edge AI applications. The EdgeTPU performs efficient on-device inference by using quantization, which reduces the computational load without compromising accuracy. This approach is particularly well-suited for tasks like image classification and object detection in real-time environments. The Google Coral USB's ability to handle quantized TensorFlow Lite models makes it an excellent choice for low-latency, high-throughput edge applications, where both performance and power efficiency are critical.


\begin{table}[h!]
\centering
\scriptsize 
\begin{tabularx}{\columnwidth}{p{2.5cm}|X}
\textbf{Model} & \textbf{Training Information} \\ \toprule

\textbf{KNN, SVM, DT, Linear Classifier} & 
\begin{tabular}[c]{@{}l@{}}
- Training Framework: TensorFlow \\
- Dataset: MNIST \\
- Training Info: Lightweight models optimized for \\low-power edge devices.
\end{tabular} \\ \hline

\textbf{ANN, CNN, FFNN, R-CNN} & 
\begin{tabular}[c]{@{}l@{}}
- Training Framework: TensorFlow Keras \\
- Dataset: MNIST \\
- Training Info: Quantization and pruning applied for \\enhanced inference speed.
\end{tabular} \\ \hline

\textbf{ResNet-50, MobileSSD} & 
\begin{tabular}[c]{@{}l@{}}
- Training Framework: TensorFlow Keras \\
- Dataset: ImageNet \\
- Training Info: Mixed precision (FP16) for improved \\performance.
\end{tabular} \\ \hline

\textbf{TinyBERT} & 
\begin{tabular}[c]{@{}l@{}}
- Training Framework: TensorFlow Keras \\
- Dataset: GLUE \\
- Training Info: Model distillation for efficient \\edge inference.
\end{tabular} \\ \hline

\textbf{Phi-2-Orange} & 
\begin{tabular}[c]{@{}l@{}}
- Training Framework: TensorFlow Keras \\
- Dataset: OpenAssistant \\
- Training Info: Optimized for performance on \\powerful edge devices.
\end{tabular} \\ \bottomrule

\end{tabularx}
\caption{Models, training tools, datasets, and training information for edge devices.}
\label{table:edge_training}
\end{table}

\subsubsection{Model Training}

The training procedures differ based on the model categories detailed below. 
Note that, vision and NLP are the key applications in edge AI, which motivated us to choose representative datasets for both applications.

\begin{itemize}
    \item \textit{KNN, SVM, DT, Linear Classifiers, ANN, CNN, FFNN, and R-CNN:} These models are trained using well-established datasets such as \textit{MNIST}~\cite{mnist_dataset} over TensorFlow. The training workflow includes feature extraction, data normalization, and hyperparameter tuning to achieve optimal accuracy. These models are utilized for multi-class classification tasks.

    \item \textit{ResNet-50 and MobileSSD:} For these models, the training involves data augmentation techniques, regularization methods like dropout, and optimization approaches such as learning rate adjustment. Training is conducted on the \textit{ImageNet}~\cite{imagenet_dataset} dataset using TensorFlow. The models are trained to perform multi-class classification.

    \item \textit{TinyBERT and Phi-2-Orange:} For large language models, TinyBERT is trained with the \textit{GLUE}~\cite{wang2019glue} dataset, which is specifically designed for question-answering tasks. The training involves fine-tuning a pre-trained BERT model by adding additional layers tailored to \textit{GLUE} tasks. This process includes fine-tuning hyperparameters such as learning rate and batch size to optimize the model's performance in question answering.

    The Phi-2-Orange model is trained on the \textit{OpenAssistant} dataset \textsc{Oasst1}~\cite{köpf2023openassistant}, which is used for intent classification in conversational contexts. The training involves processing labeled prompts that represent various user intents (\eg{} information requests, questions, complaints). The model is fine-tuned through transfer learning techniques to improve classification accuracy on this dataset. 
\end{itemize}

The training process of the above models is summarized in Table~\ref{table:edge_training}. All trained models are saved in the H5 file format to retain their weights and architecture in the TensorFlow Lite formats. This standard deployment strategy also ensures fair and consistent assessments of inference in the edge. 

\subsection{Selecting Tools to Develop a Measurement Scheme (RQ2-RQ3)} \label{measurement-tool}

This section elaborates on the specific selection of tools and techniques for uniform and fair performance and resource usage measurements. We also outline a measurement scheme for edge AI. 



\textbf{Tools Selection:} It is essential to select tools that are usable across different edge devices and frameworks. Alongside, the selected tools should not require any modifications in the applications. Thus, we carefully choose tools for consistent and fair measurements. We collect the inference results of TensorFlow Lite, \textit{OpenVINO}, or \textit{TensorRT} for a chosen model to derive the accuracy and the F1 score using ground truths, as these frameworks do not offer any tools to directly measure performance metrics. 

In the case of time measurements, we use the Python \texttt{time}\footnote{https://docs.python.org/3/library/time.html} module due to its simplicity, precision, and inclusion in Python's standard library. Specifically, we log the start and finish time using \texttt{time.strftime()} of the inference to later process the logged information to calculate the time. 
We use the \texttt{psutil}\footnote{https://pypi.org/project/psutil/} library to measure the memory utilization (total RAM usage) of the inference. It is a standard Python library that offers interfaces to measure memory usage at different granularity, e.g., at the process or system level.  

The power consumption measurement is the most challenging task due to the lack of software tools for various edge devices. As indicated in Table~\ref{table:energy_measurement_tools}, no single software solution ensures compatibility with all platforms. Thus, we adopt a hardware-based approach using \textit{USB power meter}\cite{powerMeter} due to its accuracy and ability to capture real-time power measurements across various edge devices. Using this power meter, we record fine-grained, time-stamped power readings throughout the inference process, enabling accurate profiling of system power consumption.


\textbf{Measurement Scheme:} To ensure consistent and fair performance and power measurements across edge devices, we design a measurement scheme that isolates the inference phase. All devices are powered using their respective official adapters and measurements are conducted under stable thermal conditions with active cooling when necessary. To ensure accurate power profiling, we remove interference from background processes, capturing only the power consumed during model inference.

The process begins with an initialization phase that captures the baseline power consumption in an idle state. This baseline is later subtracted to isolate the inference-specific power consumption. Following initialization, the measurement scheme triggers inference execution while logging time, memory usage, and prediction outputs. Concurrently, a USB power meter collects power samples at fixed intervals, which are recorded via device APIs. These logs are subsequently processed to compute metrics such as inference time, memory utilization, and power consumption. The detailed implementation of the proposed scheme is presented in the next section.

\begin{table}[h!]
\centering
\renewcommand{\arraystretch}{1.2} 
\setlength{\tabcolsep}{3pt} 
\scriptsize 
\begin{tabular}{l|l|l}
\textbf{Device} & \textbf{Tool Type} & \textbf{Tool Name} \\ \toprule

\multirow{4}{*}{\textbf{RPi}} 
& Software & PiJuice API \\ \cline{2-3} 
& Software & PowerAPI \\ \cline{2-3} 
& Hardware & USB Power Meter \\ \cline{2-3} 
& Hardware & INA219/INA226 Power Monitor \\ \midrule

\multirow{3}{*}{\textbf{INCS}} 
& Software & Intel® Power Gadget \\ \cline{2-3} 
& Software & psutil \\ \cline{2-3} 
& Hardware & USB Power Meter \\ \midrule

\multirow{3}{*}{\textbf{NJn}} 
& Software & tegrastats \\ \cline{2-3} 
& Software & Jetson Power Monitor \\ \cline{2-3} 
& Hardware & USB Power Meter \\ \midrule

\multirow{3}{*}{\textbf{GCU}} 
& Software & PowerAPI \\ \cline{2-3} 
& Software & psutil \\ \cline{2-3} 
& Hardware & USB Power Meter \\ \bottomrule

\end{tabular}
\caption{Energy consumption measurement tools for RPi, INCS, NJn, and GCU.}
\label{table:energy_measurement_tools}
\end{table}

%% file: 05_implementation.tex
This section details the deployment workflow and measurement setup for evaluating AI models on edge devices. We describe the standardized procedure used to deploy models, execute inferences, and collect performance and energy metrics across all platforms.

\subsection{Device Setup}

 We follow a standardized setup process tailored for five distinct hardware configurations: (i) Raspberry Pi (RPi), (ii) Raspberry Pi with Intel Neural Compute Stick (RPi+INCS), (iii) Raspberry Pi with Google Coral USB Accelerator (RPi+GCU), (iv) NVIDIA Jetson Nano using LiteRT (NJn-LiteRT), and (v) NVIDIA Jetson Nano with TensorRT optimizations (NJn-TensorRT). The standard process prepares each edge device variant for inference execution and measurement in a fair and consistent manner. 



We begin by flashing the latest RPi operating system (64-bit) to run the LiteRT \cite{TensorFlowLite} model inference. After configuring system dependencies, we install Python packages such as \texttt{tflite-runtime}, \texttt{psutil}, and \texttt{python3-pip} within a dedicated virtual environment to isolate our environment and maintain reproducibility. This serves as the baseline device for lightweight model inference with any RPi device.
In the case of \textsc{RPi + INCS}, we additionally install OpenVINO \cite{IR} to enable accelerated inference using Intel's Myriad X Vision Processing Unit.  
We install \texttt{tflite-runtime} along with the \texttt{libedgetpu1-std} driver, which enables communication with the EdgeTPU hardware for \textsc{RPi+GCU}. We also use the \texttt{edgetpu-compiler} to convert LiteRT models into an EdgeTPU-compatible format, allowing efficient inference acceleration. This tool-chain ensures that Coral executes quantized models optimized for its architecture. 

We consider two setups for Nvidia Jetson Nano with LiteRT (\textsc{NJn-LiteRT}) and TensorRT (\textsc{NJn-TensorRT}). In the former case, we load the NJn with the standard LiteRT runtime with packages such as \texttt{python-pip3}, \texttt{tflite\-runtime}, and \texttt{psutil}. This baseline setup offers inference performance without TensorRT optimizations, which is the case in the \textsc{NJn-TensorRT}. It leverages NVIDIA’s TensorRT SDK for layer fusion, quantization, and GPU-accelerated inference to maximize speed and efficiency. 
In all configurations, trained models are first saved in H5 format \cite{H5} and then converted to \texttt{.tflite}. Platform-specific conversions are applied to ensure compatibility and performance.


\subsection{Inference Measurement}

\begin{figure}[!t]
\centering
\includegraphics[width=.9\columnwidth, height=.8\textheight, keepaspectratio]{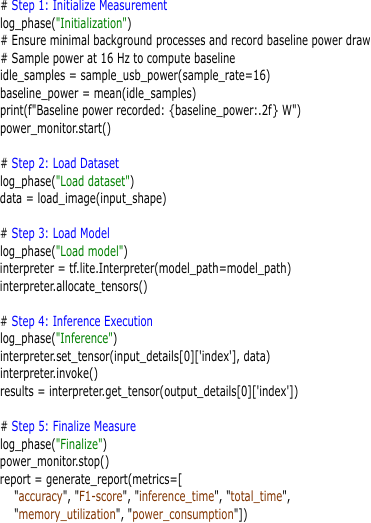}
\caption{Unified inference measurement script for edge AI evaluation.}
\label{fig:measurement_script}
\end{figure}

We developed a Python-based measurement script following the measurement steps depicted in Section~\ref{measurement-tool}. Thus, we deploy this script (see Figure~\ref{fig:measurement_script}) in each of the edge devices, where the measurement process consists of five sequential steps: (i) record the baseline power consumption in a near-idle state over a 3-second interval, corresponding to the first 48 power samples collected by the USB meter, (ii) load the test dataset, (iii) load the model, (iv) execute inference, and (v) log performance metrics. 

The LiteRT interpreter is used for inference on RPi and GCU; GCU also uses the EdgeTPU runtime to offload inference operations to its dedicated ASIC. On INCS, OpenVINO's Inference Engine executes the IR-optimized models. In NJn, inference is conducted by utilizing LiteRT (default) or TensorRT (optimized) models. This uniform process across all devices ensures fair comparisons in evaluating inference performance and resource consumption, maintaining consistency in measurement and reproducibility of results.

We publicly release our entire measurement scheme and associated scripts as an open source repository to support reproducibility and facilitate broader adoption \cite{edgeAI-gitRepo}. The repository includes all model preparation pipelines, device-specific deployment scripts, and the unified measurement script used across devices. Users can adapt the scripts for their own inference tasks on edge platforms while ensuring consistent measurement of performance and energy metrics.

\subsection{Evaluation Metrics}


We consider the following evaluation metrics to evaluate the proposed measurement scheme, where each test is run multiple times to get the average and 95\% confidence interval. 

\noindent
\textbf{F1-score:}
The F1-score, defined as the harmonic mean of precision and recall, offers a balanced measure of a model’s classification performance.
A high F1-score indicates strong performance in correctly identifying relevant instances while minimizing false positives and false negatives. F1-score is reported as percentages (\%).

\noindent
\textbf{Inference Time:}
Inference time measures the duration taken by a model to complete an inference task. 
Inference time is typically measured by recording timestamps before and after each operation and calculating the time differences (in seconds).

\noindent
\textbf{Memory Utilization:}
Memory utilization refers to the total memory consumed during inference, which includes memory consumption by the attached devices (\eg{} GPU memory on the NVIDIA Jetson Nano) and the host machine (\ie{} system RAM utilized by the CPU). In our evaluation, memory utilization is reported in MegaBytes (MB).

\noindent
\textbf{Power Consumption:}
Inference power consumption refers to the total power (in watts) consumed per input sample during the inference phase. We use a USB power meter to log power readings at fixed intervals (16 Hz sampling) throughout the inference process. The total power is then computed by summing the recorded values over the entire inference duration.

%% file: 06_evaluation.tex
This section presents the evaluation results by answering the following research questions.

\begin{description}
    \item[\textbf{RQ1.}] 
    \textit{Does domain-specific hardware and software design boost edge AI?}


    \item[\textbf{RQ2.}]
    \textit{What are the trade-offs between performance and resource usage in edge AI deployments?}

    \item[\textbf{RQ3.}]
    \textit{Which parameters can further be tuned for better power consumption in deep and large models?}
\end{description}

\subsection{Does domain-specific hardware and software design boost edge AI? (RQ1)}

This research question seeks to investigate the performance and resource utilization trends of different edge devices and identify their similarities and differences. Specifically, we assess the impact of various hardware platforms and their lightweight frameworks on performance and resource usage in edge AI, aiming to provide insights into the suitability of particular hardware and frameworks for edge applications.

\begin{figure*}[htbp]
\centering
\includegraphics[width=\textwidth]{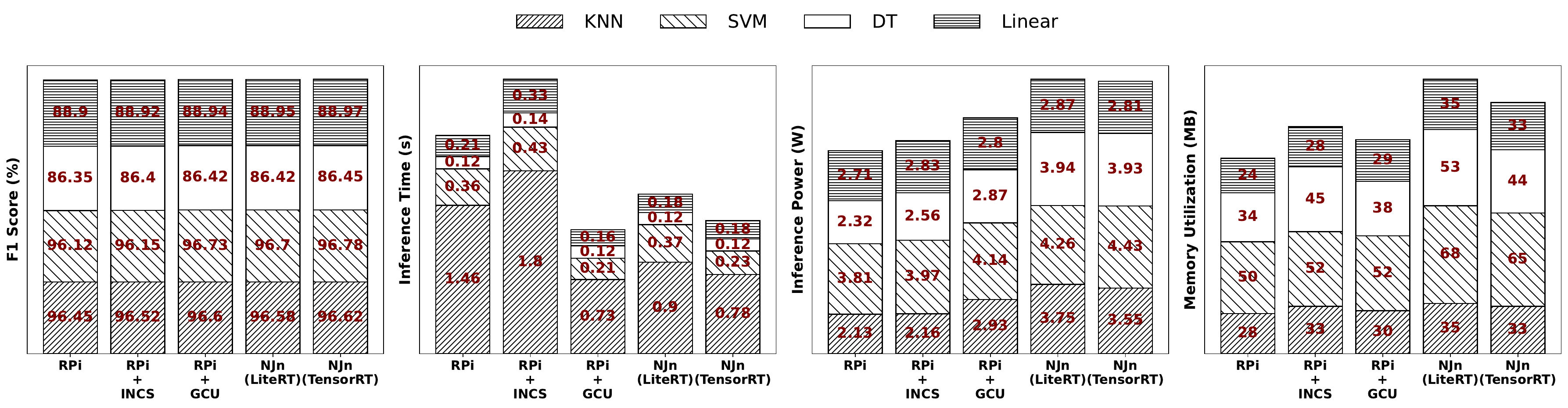}
\\
\hspace{2.5em}(a) F1-score\hspace{6.5em}(b) Inference time\hspace{5em}(c) Inference power\hspace{3em}(d) Memory utilization
\caption{Evaluation results of traditional machine learning models across edge devices.}
\label{fig:ml_figs}
\end{figure*}

\textbf{Traditional ML Models:}
Figure \ref{fig:ml_figs} presents the results for the traditional machine learning models. 
From the performance (\textit{i.e.,} \textit{F1-score}) perspective, expectedly, all the devices perform almost equally well.
NVIDIA Jetson Nano and Google Coral surpass Raspberry Pi and Intel Neural Compute  Stick setups in terms of \textit{inference time} due to their advanced hardware capabilities, such as a GPU that accelerates parallel computations and the EdgeTPU that is optimized for inference.
From the lens of \textit{inference power} consumption, Raspberry Pi shows better efficiency than other devices, which is a single-board ARM processor-based embedded system usually require low power.
Finally, these edge devices exhibit a similar memory usage trend while Nvidia Jetson Nano is a slightly higher consumer. 

We observe two groups of ML models: SVM and KNN outperform DT and regression in \textit{F1-score} as expected. However, the trend completely reverts in the case of \textit{inference time} as SVM and KNN involve intensive mathematical operations. 
Although inference power is typically expected to mirror inference time, we notice some mismatches, particularly for CPU-bound models, due to device-specific characteristics such as fixed power overheads (\textit{i.e.,} baseline power draw unrelated to workload) and limited dynamic scaling (\textit{i.e.,} inability to adjust power use in response to computational demand). This suggests that energy readings may not always scale linearly with runtime, especially for lightweight tasks.
Simpler models like DT and regression do not gain significant benefits from hardware acceleration, allowing the Raspberry Pi to perform relatively well. 

\begin{figure*}[htbp]
\centering
\includegraphics[width=\textwidth]{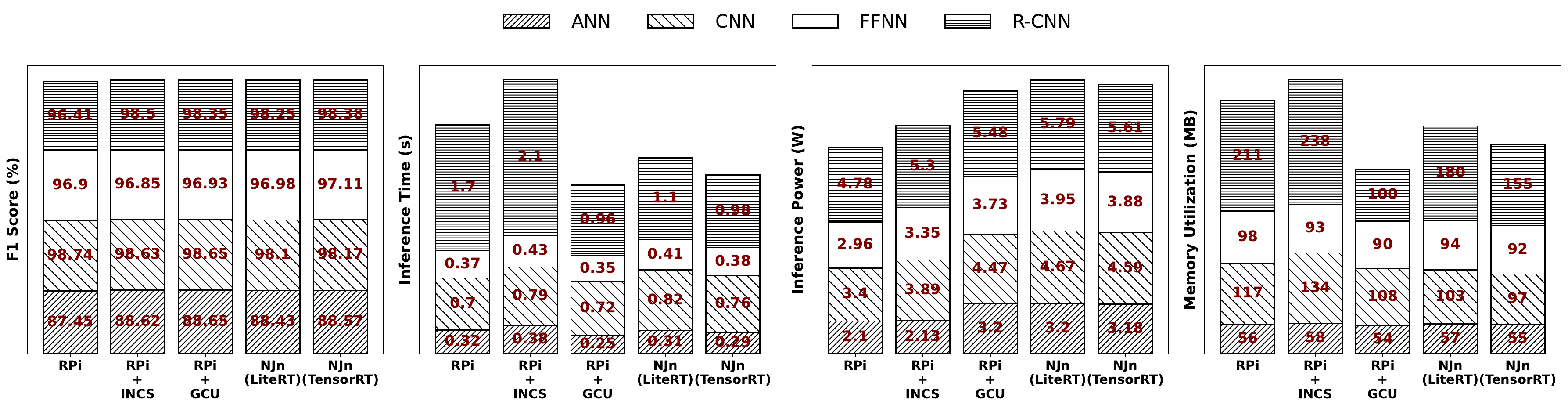}
\\
\hspace{2.5em}(a) F1-score\hspace{6.5em}(b) Inference time\hspace{5em}(c) Inference power\hspace{3em}(d) Memory utilization
\caption{Evaluation results of neural network models across edge devices.}
\label{fig:NN_figs}
\end{figure*}

\textbf{Neural Network Models:}
The performance of these models is presented in Figure~\ref{fig:NN_figs}, where the trend is similar as in traditional models. For instance, all devices have a similar F1-score. In the case of inference time, Google Coral and Jetson Nano show better inference time and memory utilization than the other devices, whereas Raspberry Pi consumes the lowest energy as expected. 

These performance variances stem from the unique hardware optimizations of each device. The Jetson Nano’s GPU and Google Coral USB's TPU excel at parallel processing, which is essential for tasks such as CNN inference, providing an edge over the Raspberry Pi’s general-purpose CPU. 
However, the Raspberry Pi’s CPU, while efficient in terms of power consumption, is less capable of handling intensive tasks, making it more suitable for lighter models like ANN and FFNN. The Movidius Myriad X VPU in Intel Neural Compute Stick, designed to accelerate inference on CPU-based systems like the Raspberry Pi, provides substantial improvements but still falls short of the Jetson Nano's GPU and Google Coral's TPU performance for complex models like R-CNN.

\begin{figure*}[htbp]
\centering
\includegraphics[width=\textwidth]{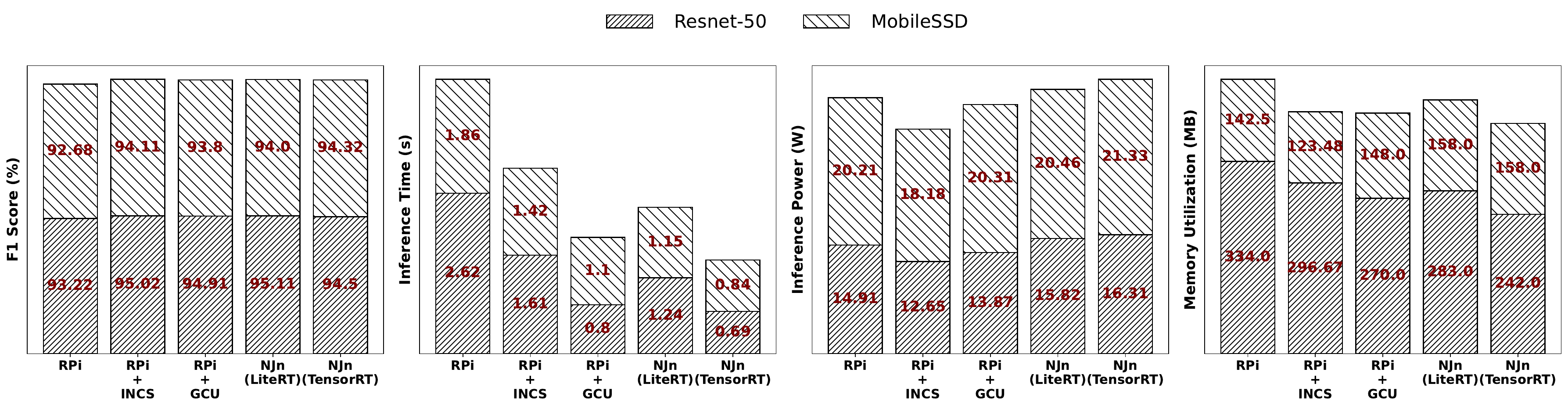}
\\
\hspace{2.5em}(a) F1-score\hspace{6.5em}(b) Inference time\hspace{5em}(c) Inference power\hspace{3em}(d) Memory utilization
\caption{Evaluation results of deep learning models across edge devices.}
\label{fig:DL_figs}
\end{figure*}

\textbf{Deep Learning Models:}
Figure~\ref{fig:DL_figs}
presents the performance of deep learning models evaluated across different hardware platforms. 
All devices perform almost equally in the case of \textit{F1-score}. Nvidia Jetson Nano with TensorRT achieves the fastest inference and lowest memory usage, with competitive energy efficiency. Intel Neural Compute Stick and Google Coral are the next two devices in terms of inference time and power, while Raspberry Pi is the slowest and consumes the highest memory. In terms of models, ResNet-50 and MobileSSD, both being resource-intensive, showcase performance disparities, especially in inference time and memory utilization.


\begin{figure*}[htbp]
\centering
\includegraphics[width=\textwidth]{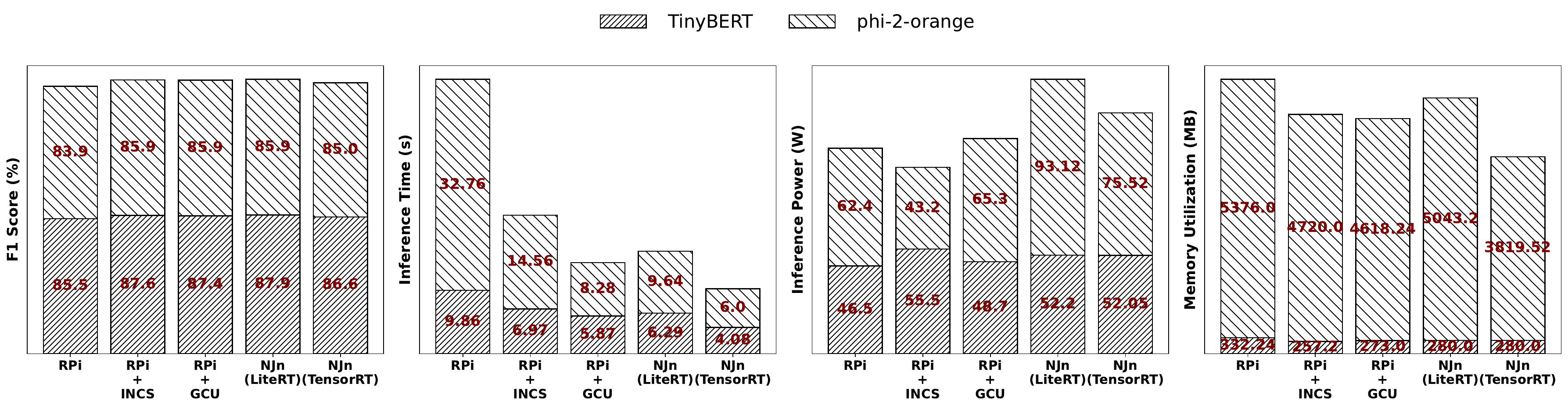}
\\
\hspace{2.5em}(a) F1-score\hspace{6.5em}(b) Inference time\hspace{5em}(c) Inference power\hspace{3em}(d) Memory utilization
\caption{Evaluation results of large language models across edge devices.}
\label{fig:LLM_figs}
\end{figure*}

\textbf{Large Language Models:}
Figure~\ref{fig:LLM_figs} presents the performance characteristics of the considered devices when experimented with Large Language Models (LLMs).
All the devices exhibit similar F1-score patterns. However, Jetson Nano (with TensorRT optimization) is the fastest devices in terms of inference time. In addition, Jetson Nano exhibits the lowest memory utilization among the considered devices.
Respberry Pi is the slowest device taking multi-fold time compared the the fastest device in terms of inference time.
It also consumes the highest memory to perform the task at hand. 
Although this performance trend is expected, the performance of the chosen LLMs show some interesting insights. 
TinyBERT outperforms Phi-2 orange, especially in inference time and memory demand, making it a suitable choice for resource-constrained edge devices. This disparity stems from TinyBERT's architecture, which is derived through model distillation to optimize for efficiency while retaining task performance \cite{jiao-etal-2020-tinybert}.

\subsubsection{How Do Light AI Frameworks Impact the Characteristics of AI Models?}
The assessment of different models across various devices highlights how LiteRT, Intermediate Representation (IR), TensorRT, and EdgeTPU frameworks offer distinct optimization strategies and performance outcomes. \textbf{LiteRT} is specifically optimized for edge devices, employing techniques such as quantization and operator fusion to enhance processing speed and efficiency~\cite{TensorFlowLite}. However, its optimizations might not fully exploit the advanced capabilities of more powerful hardware like the NVIDIA Jetson Nano. \textbf{TensorRT} is tailored for NVIDIA GPUs~\cite{TensorRT} and features sophisticated optimizations, including layer fusion and precision calibration. These advanced techniques significantly boost inference speed, accuracy, and overall efficiency, particularly on NVIDIA Jetson Nano devices. For instance, deep learning models such as Resnet-50 and MobileSSD see substantial performance improvements when deployed on the NVIDIA Jetson Nano with TensorRT.

\textbf{Intermediate Representation (IR)}, utilized with OpenVINO, provides a versatile hardware abstraction layer that supports customized optimizations~\cite{IR}. While IR can deliver improvements in performance, it often results in less pronounced gains compared to TensorRT. On the other hand, Google Coral USB EdgeTPU, used with frameworks like LiteRT, offers specialized hardware acceleration for edge devices, significantly improving inference speed and power consumption. For example, when deployed with the Raspberry Pi, models like TinyBERT experience noticeable performance improvements in both inference time and power consumption compared to configurations using the Intel Neural Compute Stick (see Figure~\ref{fig:LLM_figs}).

\findingsbox{
All devices offer similar F1-score while their performance varies in other metrics.
Jetson Nano and Google Coral are the fastest devices irrespective of the chosen models, where the inference time is proportional to the complexity of the model architectures. This trend reverts in the case of inference power; however, the difference is not significant, which infer that specifically designed edge devices are good fit for complex models to balance between performance and power consumption. 
A hardware-agnostic framework can offer consistency and flexibility in implementation while preserving optimal performance. However, resource-constrained devices (e.g., RPi) without such frameworks can leverage IR and EdgeTPU to improve performance,  where the choice between these two depends on AI models.}

\subsection{What are the trade-offs between performance and resource usage? (RQ2)}

\begin{table*}[h!]
\centering
\caption{Performance comparison across hardware targets. Symbols represent hardware targets: $\diamond$ (Raspberry Pi), $\odot$ (Neural Stick), $\Delta$ (Google Coral), $\star$ (Jetson Nano), and $\sim$ (Tie). A tie ($\sim$) indicates that two or more hardware targets show comparable performance.}
\begin{tabular}{|l|c|c|c|c|}
\hline
\textbf{Model} & \textbf{F1 Score ($\uparrow$)} & \textbf{Inference Time ($\downarrow$)} & \textbf{Inference Power ($\downarrow$)} & \textbf{Memory Utilization ($\downarrow$)} \\
\hline
KNN & $\star$ & $\Delta$ & $\diamond$ & $\diamond$ \\
SVM & $\star$ & $\Delta$ & $\diamond$ & $\diamond$ \\
DT & $\sim$ & $\sim$ & $\diamond$ & $\sim$ \\
Linear Regression & $\sim$ & $\Delta$ & $\diamond$ & $\diamond$ \\
ANN & $\Delta$ & $\star$ & $\diamond$ & $\diamond$ \\
CNN & $\Delta$ & $\star$ & $\diamond$ & $\diamond$ \\
FFNN & $\star$ & $\star$ & $\diamond$ & $\star$ \\
R-CNN & $\star$ & $\star$ & $\diamond$ & $\star$ \\
ResNet-50 & $\star$ & $\Delta$ & $\odot$ & $\star$ \\
MobileSSD & $\star$ & $\Delta$ & $\odot$ & $\diamond$ \\
TinyBERT & $\star$ & $\star$ & $\diamond$ & $\star$ \\
phi-2-orange & $\star$ & $\star$ & $\diamond$ & $\diamond$ \\
\hline
\end{tabular}
\label{tab:performance_comparison}
\end{table*}

Table~\ref{tab:performance_comparison} presents a comparative analysis of performance across different edge devices for all models. Traditional models such as KNN, SVM, DT, and Linear Regression are suitable for deployment on a Raspberry Pi when performance requirements are modest; however,  Jetson Nano is preferable for more demanding applications. Jetson Nano excels with neural network models due to its specialized design, and TensorRT optimizations further enhance its capabilities.
Consequently, models such as ResNet-50 and MobileSSD achieve superior performance, particularly in terms of F1-score and inference time, on Jetson Nano.


The Google Coral EdgeTPU provides a notable performance enhancement for large language models like TinyBERT and phi-2-orange on the Raspberry Pi, offering improved inference speed and power consumption compared to the Intel Neural Compute Stick. This highlights the role of EdgeTPU in balancing performance with resource consumption in lower-end devices. TinyBERT, in particular, benefits significantly from EdgeTPU, showing better overall efficiency than with the Neural Compute Stick. Conversely, while TinyBERT and phi-2-orange exhibit optimal results on the Jetson Nano, their performance on the Raspberry Pi with EdgeTPU and other hardware varies, suggesting that choosing the right optimization and hardware depends on the specific application requirements. 

In summary, while F1-scores remain consistent across devices, there is noticeable variation in inference time, inference power, and memory efficiency. Jetson Nano offers the best overall performance, benefiting from GPU acceleration and TensorRT. Raspberry Pi is energy-efficient but lags in speed and memory handling. Coral and Intel NCS provide balanced performance, with Coral excelling in speed and NCS in power consumption for specific models. For large language models, Jetson Nano supports longer inputs efficiently, whereas Raspberry Pi and NCS handle only shorter sequences. TinyBERT outperforms phi-2-orange in power and memory efficiency, making it preferable for constrained deployments.


\subsection{Which parameters can further be tuned for better energy usage in deep and large models? (RQ3)}

The above evaluation results confirm the suitability of edge AI for various applications like autonomous vehicles. These applications may require a certain response time and inference accuracy. Hardware- and framework-level optimizations (e.g., parallelization, pruning, and quantization) play a significant role in meeting these requirements. In addition, tuning external parameters during inference can further enhance performance and optimize resource utilization. This section examines how deep and large models can benefit from such parameter tuning in edge deployments. Note that, traditional and simpler neural network models are excluded, as they already have low resource demands. We also excluded Jetson Nano with LiteRT as TensorRT offers the best performance. 

\subsubsection{Parameter tuning for deep neural networks}
One of the primary application domains for the selected deep neural network models is image processing. Accordingly, our evaluations are based on representative image data (see Section~\ref{subsec:AI framework selection}). In such applications, two key inference parameters, image resolution (input size) and the number of images processed per task (i.e., batch size), can significantly affect both model performance and resource consumption. 

\textbf{Input Size Impact:}
In vision tasks, input size (image resolution) directly affects the amount of computation required during inference. Larger images encode more features but increase the number of operations in convolutional layers, leading to longer inference times and higher energy use. The extent of this impact varies depending on the model architecture and the hardware used for deployment. We examine this trade-off using two input sizes: 224x224 (standard resolution utilized in ResNet-50) / 300x300 (standard resolution utilized in MobileSSD) and 512x512 (higher detail). 

\begin{figure*}[htbp]
\centering
\includegraphics[width=\textwidth]{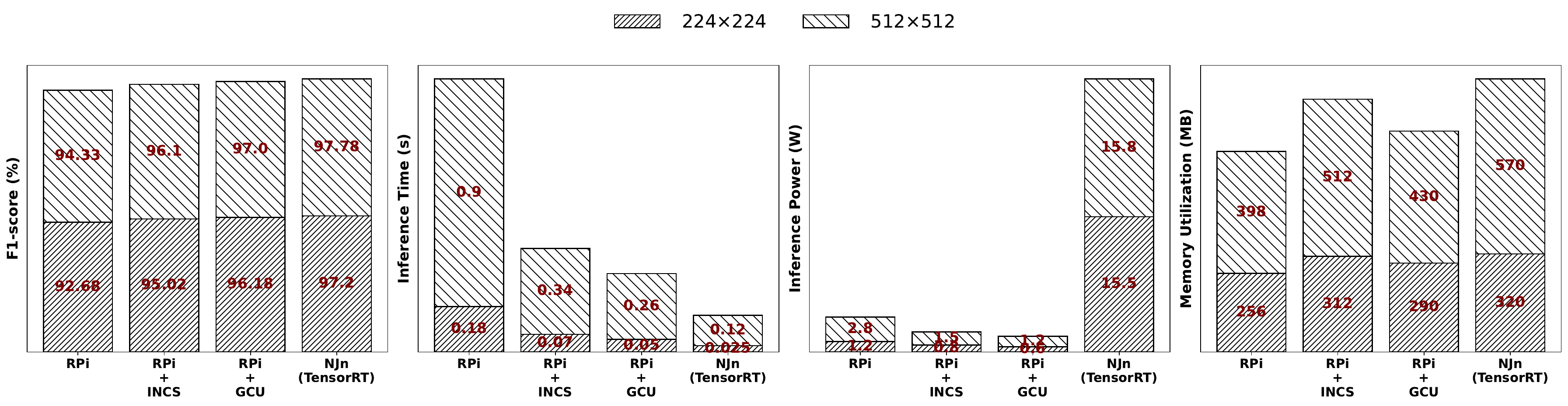}
\\
\hspace{2.5em}(a) F1-score\hspace{6.5em}(b) Inference time\hspace{5em}(c) Inference power\hspace{3em}(d) Memory utilization
\caption{The impact of input size on ResNet-50 performance across edge devices.}
\label{fig:Resnet_input_size_figs}
\end{figure*}

Figure~\ref{fig:Resnet_input_size_figs} illustrates how increasing image size impacts the performance of ResNet-50. Jetson Nano maintains inference time and low power with rising resolution while maintaining superior F1-score due to its GPU and TensorRT optimizations. In contrast, Raspberry Pi exhibits drastic increases in both inference time and power consumption. Coral and INCS show moderate increases in resource usage, but their limited hardware capabilities lead to a twofold rise in power consumption at higher resolution. These findings are aligned with Qi \etal{}~\cite{Qi2018}, who show that input size optimization is crucial for real-time deployment of CNNs in edge AI.

\begin{figure*}[htbp]
\centering
\includegraphics[width=\textwidth]{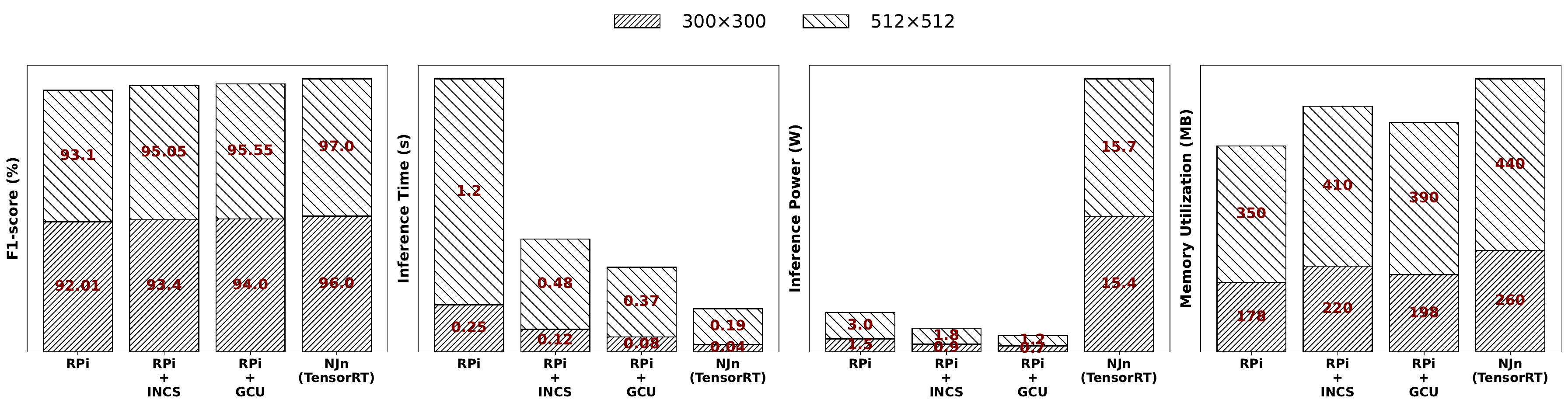}
\\
\hspace{2.5em}(a) F1-score\hspace{6.5em}(b) Inference time\hspace{5em}(c) Inference power\hspace{3em}(d) Memory utilization
\caption{The impact of input size on MobileSSD performance across edge devices.}
\label{fig:MobileSSD_input_size_figs}
\end{figure*}

For MobileSSD (Figure~\ref{fig:MobileSSD_input_size_figs}), which is optimized for real-time object detection, performance improves at higher resolutions, benefiting from enhanced object localization due to increased input detail. Jetson Nano continues to lead in inference speed and power efficiency, while Raspberry Pi exhibits high inference time. Comparing the two deep learning models on Jetson Nano, ResNet-50 achieves higher F1-score and faster inference than MobileSSD, while remaining competitive in power consumption. These results highlight the importance of selecting both task-specific models and appropriate hardware for edge AI deployments.

\textbf{Batch Size Impact:}
Batch size determines how many input samples are processed in parallel during inference. While large batch sizes improve throughput, they can lead to memory saturation and increased latency on memory-constrained devices. On the contrary, smaller batch sizes may result in under-utilization. We observe that both ResNet-50 and MobileSSD share similar batch size trends across devices, allowing us to report a unified summary.
\begin{itemize}
    \item Raspberry Pi: Batch size 1 is the only viable option due to severe memory constraints; higher values result in paging and degraded performance.
    \item Intel NCS: Batch sizes between 4 and 8 yield improved throughput and power efficiency without memory overload.
    \item Google Coral TPU: Performs best with batch sizes in the 8–16 range, balancing inference time and power consumption.
    \item Jetson Nano: Supports batch sizes of 16-32 effectively, leading to up to 20\% lower per-sample inference time while keeping power consumption reasonable.
\end{itemize}

This consistency across both models suggests that optimal batch size tuning is primarily governed by device memory bandwidth and processing capabilities rather than specific model architecture. Thus, identifying device-specific batch ranges enables generalized tuning strategies for deep neural network models on edge hardware.

\subsubsection{Parameter tuning for large language models}

To support natural language applications on the edge, such as voice assistants, on-device translation, or real-time summarization, large language models (LLMs) must be adapted to operate efficiently under constrained resource budgets. Unlike deep neural networks used in vision tasks, LLMs process tokenized text sequences, making the number of tokens and token window size the dominant inference parameters \cite{chitty2024llm}. Optimizing these parameters is essential to control inference time, memory usage, and power consumption while maintaining acceptable performance in edge environment \cite{arya2025understanding}. 

\textbf{Input Token Length Impact:} 
Input token length refers to the number of tokens fed into the model during each inference request. Longer inputs can improve contextual understanding but increase the computational load and memory footprint. As shown in Table~\ref{tab:input_size_experiment}, Raspberry Pi and Intel NCS exhibit stable performance with 500-1000 tokens. Beyond this range, inference becomes increasingly costly. Jetson Nano demonstrates the highest tolerance, processing up to 2048 tokens efficiently with acceptable power overhead. Among the tested models, TinyBERT generally shows more stable performance than phi-2-orange, consuming less memory and power.

\textbf{Token Window Size Impact:} 
Token window length affects the model's ability to retain context during generation. Larger windows improve language coherence but amplify inference time and power consumption. Table~\ref{tab:token_window_experiment} demonstrates that both models achieve reasonable performance with a 512-token window on Raspberry Pi and a 1024-token window on NCS. Jetson Nano can handle up to 2048 tokens with efficient memory and time trade-offs. Similar to input token length, TinyBERT proves more resource-friendly across devices.


\begin{table*}[htbp]
  \centering
  \caption{Impact of input token length on LLM performance. Metrics are reported along with their confidence interval in square brackets.}
  \label{tab:input_size_experiment}
  \resizebox{\textwidth}{!}{%
  \begin{tabular}{p{3cm}lcccccc}
    \toprule
    \textbf{Device} & 
    \textbf{Model} & 
    \textbf{Optimal Input Size (tokens)} & 
    \textbf{F1-score (\%)} & 
    \textbf{Inference Time (s)} & 
    \textbf{Inference Power (W)} & 
    \textbf{Memory Utilization (MB)} \\ 
    \midrule
    \multirow{2}{*}{Raspberry} & 
    TinyBERT & 500-1000 
    & 85.50 [84.12, 86.88] & 9.86 [9.35, 10.37] & 46.5 [34.05, 51.45] & 332.24 \\
    & phi-2-orange & 500-1000 
    & 83.90 [82.78, 84.83] & 32.76 [29.48, 36] & 62.4 [48.96, 75.84] & 5,250 \\
    \midrule
    \multirow{2}{*}{Raspberry Pi + Stick} & 
    TinyBERT & 1024 
    & 87.60 [86.32, 88.88] & 6.97 [6.46, 7.48] & 43.2 [29.76, 56.64] & 257.2 \\
    & phi-2-orange & 1024 
    & 85.90 [84.77, 86.98] & 14.56 [11.64, 17.47] & 55.5 [42.45, 58.65] & 4,620 \\
    \midrule
    \multirow{2}{*}{NVIDIA Jetson Nano} & 
    TinyBERT & 2048 
    & 87.90 [86.70, 89.10] & 6.29 [5.78, 6.8] & 52.2 [40.05, 64.35] & 280 \\
    & phi-2-orange & 2048 
    & 85.90 [84.75, 87.06] & 9.64 [8.37, 10.92] & 93.12 [79.52, 106.72] & 4,930 \\
    \bottomrule
  \end{tabular}%
  }
\end{table*}

\begin{table*}[htbp]
  \centering
  \caption{Impact of token window size on LLM performance. Metrics are reported along with their confidence interval in square brackets.}
  \label{tab:token_window_experiment}
  \resizebox{\textwidth}{!}{%
  \begin{tabular}{p{3cm}lcccccc}
    \toprule
    \textbf{Device} & 
    \textbf{Model} & 
    \textbf{Token Window Size} & 
    \textbf{F1-score (\%)} & 
    \textbf{Inference Time (s)} & 
    \textbf{Inference Power (W)} & 
    \textbf{Memory Utilization (MB)} \\ 
    \midrule
    \multirow{2}{*}{Raspberry} & 
    TinyBERT & 512 
    & 85.10 [83.66, 86.54] & 9.42 [8.90, 9.94] & 45.3 [32.78, 49.82] & 320.14 \\
    & phi-2-orange & 512 
    & 83.40 [82.22, 84.58] & 30.24 [27.96, 32.52] & 60.8 [46.78, 74.82] & 5,100 \\
    \midrule
    \multirow{2}{*}{Raspberry Pi + Stick} & 
    TinyBERT & 1024 
    & 87.00 [85.64, 88.36] & 6.32 [5.84, 6.80] & 41.0 [27.72, 54.28] & 249.8 \\
    & phi-2-orange & 1024 
    & 85.50 [84.32, 86.68] & 13.26 [10.90, 15.62] & 53.1 [40.22, 56.98] & 4,550 \\
    \midrule
    \multirow{2}{*}{NVIDIA Jetson Nano} & 
    TinyBERT & 2048 
    & 87.50 [86.20, 88.80] & 5.95 [5.45, 6.45] & 51.0 [38.55, 63.45] & 275 \\
    & phi-2-orange & 2048 
    & 85.70 [84.55, 86.85] & 9.10 [7.98, 10.22] & 91.2 [77.45, 104.95] & 4,890 \\
    \bottomrule
  \end{tabular}%
  }
\end{table*}

\findingsbox{
Efficient deployment of deep and language models on edge devices requires careful tuning of inference parameters. For vision tasks, larger input sizes improve F1-score but significantly raise inference time and power consumption, particularly on memory-constrained platforms like Raspberry Pi. Jetson Nano maintains stable performance even at higher resolutions due to GPU acceleration. Similarly, optimal batch sizes vary by device, Jetson Nano performs best with 16-32, Coral TPU with 8-16, and Intel NCS with 4-8. \\
For LLMs, inference efficiency is highly sensitive to input token length and window size. While Jetson Nano handles longer contexts (up to 2048 tokens), Raspberry Pi and NCS are limited to shorter sequences (500-1000 tokens). Across all configurations, TinyBERT consistently outperforms Phi-2-orange in memory and power consumption, making it preferable for constrained deployments. These insights underscore the importance of model-aware and hardware-specific parameter tuning to ensure responsive and sustainable edge inference.
}

%% file: 07_discussion.tex
\section{Threat to Validity and Discussion}

Throughout this study, we encountered several challenges that required innovative problem-solving and adaptation. 

\textbf{Construct validity:} It refers to how well the experiment actually measures what it intends to measure.
In our context, establishing a consistent method for measuring energy consumption proved to be a significant hurdle. We needed a solution that would provide reliable measurements across all devices in our study while ensuring accuracy, ease of use, and repeatability. 
There are software tools and libraries available for traditional computing system to measure consumed power. However, the absence of a software that works in our specific experimental edge environment added complexity to the process.
Additionally, with the adopted USB power meter device, we faced challenges due to the lack of documentation, which required extensive experimentation and troubleshooting to properly interface with these devices and interpret the data they provide. 
However, we mitigated these challenges by adopting a widely used USB power meter and developed scripts to sync the execution of the experiments with our energy computation logic.
In addition, we adopted established experimental practices such as conducting multiple runs per configuration and incorporating cooling intervals between successive tests to ensure accurate and consistent energy measurements.

\textbf{External validity:} It refers to the extent to which the results of the study can be generalized beyond the specific experimental context. To mitigate this threat, we selected a heterogeneous set of resource-constrained edge devices representing a wide spectrum of computational capabilities. Additionally, we evaluated a diverse suite of downstream tasks spanning traditional ML, deep learning, and natural language processing workloads to enhance the applicability and generalizability of our findings across real-world edge AI scenarios.

\textbf{Platform and model validity:} In addition to these challenges, we also encountered difficulties due to the resource constraints of the devices involved. Deploying deep learning and machine learning models posed a significant hurdle, as the devices had limited resources. Finding a suitable implementation for each model and optimizing the transformation process became imperative. Moreover, learning different methods of transforming and executing models on devices such as the Nvidia Jetson and Neural Stick demanded a considerable amount of time and effort. The search for a suitable large language model that could be effectively executed on these devices added to the complexity. Overcoming these challenges required careful consideration of resource allocation, optimization techniques, and exploration of various model implementations.

\textbf{Discussion on Future Work:} 
This section presents possible extensions of current work in the following avenues.

\textit{Automated measurement tool for edge AI:} 
We have developed a five-phase measurement scheme that can be extended and reused for measurement in edge AI. However, we manually provide relevant input to the tool for collecting and processing the outcomes. It would be beneficial to have a fully automated measurement tool that would get user intent (e.g., applications, QoS metrics) to automatically select right set of devices and models along with optimal parameter settings to have an end-to-end automated measurement system. We plan to extend our current work to develop such a system.

\textit{Automated parameter tuning:}
The evaluation results indicate that the optimal settings for different parameters vary widely depending on the specific edge devices and models. For example, lower-powered devices require smaller input sizes with modest batch sizes to balance energy consumption with latency, whereas more capable devices can efficiently process larger inputs and support larger batch sizes. Thus, it would be beneficial to have an automated parameter tuning system of the likes of May \textit{et al.} \cite{may2024dynasplit}. However, this system relies on a heuristic for parameter tuning on a single device that may not work across a diverse set of devices and models with varying performance demand. Thus, we will develop a learning-based system for automated parameter tuning. We will also test this system with representative workloads for edge AI applications.

%% file: 08_conclusion.tex
This paper rigorously evaluated the performance of various machine learning, deep learning, and large language models on different edge devices using frameworks such as LiteRT, TensorRT. Furthermore, we developed a measurement scheme for consistent and fair measurement across these models and devices.
Our results indicate that the choice of hardware and model architecture significantly impacts performance metrics such as inference time, F1 score, and power consumption. The Nvidia Jetson Nano, particularly with TensorRT optimizations, emerged as the most efficient platform for edge AI applications, offering superior performance in terms of both speed and power consumption. This makes it an excellent choice for deploying complex deep learning models in edge environments. Future work could investigate the integration of other hardware accelerators and the potential for further optimization of model architectures to enhance edge AI performance.